# Utilizing GANs for Fraud Detection: Model Training with Synthetic Transaction Data

**Mengran Zhu[1,*]**

Computer Engineering,Miami University,Oxford, OH USA
mengran.zhu0504@gmail.com

**Yulu Gong[2]**

Computer & Information Technology,Northern Arizona University,Flagstaff, AZ, USA
yg486@nau.edu

**Yafei Xiang[3]**

Computer Science,Northeastern University ,Boston, MA, USA
xiang.yaf@northeastern.edu

**Hanyi Yu[4]**

Computer Science,University of Southern California,Los Angeles, CA, USA
hanyiyu@usc.edu

**Shuning Huo[5]**

Statistics,Virginia Tech,Blacksburg, VA, USA
shuni93@vt.edu

**Abstract.** Anomaly detection is a critical challenge across various research domains, aiming to identify instances that deviate from normal data distributions. This paper explores the application of Generative Adversarial Networks (GANs) in fraud detection, comparing their advantages with traditional methods. GANs, a type of Artificial Neural Network (ANN), have shown promise in modeling complex data distributions, making them effective tools for anomaly detection. The paper systematically describes the principles of GANs and their derivative models, emphasizing their application in fraud detection across different datasets. And by building a collection of adversarial verification graphs, we will effectively prevent fraud caused by bots or automated systems and ensure that the users in the transaction are real. The objective of the experiment is to design and implement a fake face verification code and fraud detection system based on Generative Adversarial network (GANs) algorithm to enhance the security of the transaction process.The study demonstrates the potential of GANs in enhancing transaction security through deep learning techniques.

## 1. Introduction

---

[1]* Corresponding author: [Mengran Zhu]. Email: [mengran.zhu0504@gmail.com].

In the field of artificial intelligence technology, the Artificial Neural Network (ANN) stands out as one of the most promising models. ANN is a mathematical model that simulates the complex information processing mechanism of the human brain's nervous system based on knowledge of network topology. It demonstrates outstanding performance in pattern recognition, intelligent perception, combinatorial optimization, and other fields[1]. For instance, in the realm of intelligent perception, Convolutional Neural Networks (CNNs) are applied to extract image features and play a crucial role in areas such as portrait recognition and target recognition for unmanned systems. In this paper, the Generative Adversarial Network (GAN) adopts ANN as the primary framework for both the discriminator and generator, along with their derived models.

An anomaly is a pattern in the data that does not adhere to the definition of normal behavior (Chandola et al., 2009). Generative adversarial networks (GANs) and adversarial training frameworks (Goodfellow et al., 2014) have been successfully applied to model the complex and high-dimensional distributions of real-world data. This property of GANs suggests that they can be successfully employed for anomaly detection, even though their applications have only recently been explored. The task of anomaly detection using GANs involves modeling normal behavior through adversarial training processes and measuring anomaly scores to detect anomalies (Schlegl et al., 2017). GANs are also widely used in common technologies such as interoperability, autonomy, secure networks, and man-machine collaboration for unmanned systems[2]. It is evident that the development of GANs has a significant impact on the intelligent processes of unmanned systems. In this paper, the basic principles and derivative models of GANs are systematically described. Subsequently, the practical application of GANs in unmanned systems is discussed from the perspectives of intelligent perception, intelligent judgment, intelligent decision-making, and human-computer interaction. Finally, the development direction of GANs is further discussed through the emerging trends in the common technologies of unmanned systems.

## 2. GANs concept and advantages

The GAN network structure is composed of a generator and a discriminator. During the training process, the generator G continuously generates forgeries, and the discriminator D identifies whether the results generated by generator G are genuine or fake[3]. The two networks confront each other: the generator G aims to produce forgeries that deceive discriminator D, while discriminator D attempts to identify the forgeries generated by generator G. This cycle repeats as the networks train each other.

### 2.1. Introduction to GANs

The full name of GAN is Generative Adversarial Network. GAN consists of two neural networks: the generator and the discriminator. Goodfellow and others proposed the GAN framework based on the idea of a zero-sum game[4]. GANs aim to construct antagonistic generators and discriminators, establishing game relationships between them based on different objectives. The discriminator judges whether the input sample is real or not, and the generator is responsible for creating false samples that interfere with the discriminator's judgment. The performance of both networks is simultaneously improved through adversarial training.

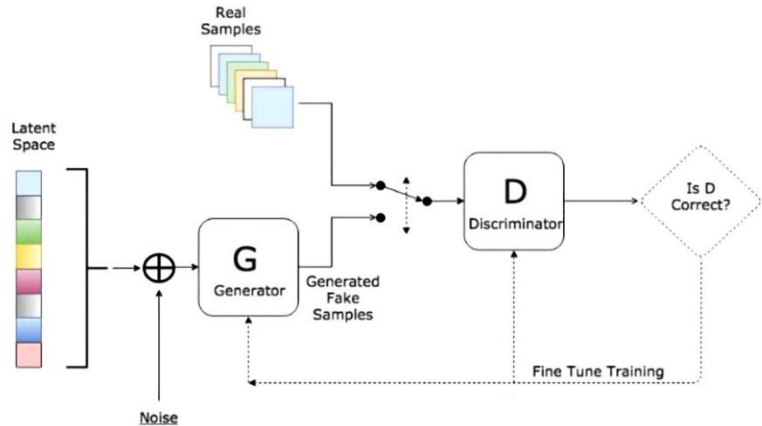

**Figure 1.** GAN principle flow chart

The main inspiration of GAN comes from the idea of zero-sum game in game theory, applied to deep learning neural networks, that is, by generating network G (Generator) and Discriminator network D (Discriminator) constantly game, and then make G learn the distribution of data **, if used in image generation, after the training is completed[5], G can generate a realistic image from a random number. The main functions of G, D are:

(1) G is a generative network that receives a random noise z (random number) through which an image is generated.

(2) D is a discriminating network that determines whether a picture is "real." Its input parameter is x, x represents a picture, the output D (x) represents the probability that x is a real picture, if it is 1, it is 100% real, and the output is 0, it is impossible to be a real picture

*2.2. GAN anomaly detection*

Anomaly detection based on GAN is a new research field. Schlegl et al.(2017), here called AnoGAN, was the first to propose this concept. Zenati et al. (2018) propose a Bigan-based approach, here called EGBAD (Efficient GAN Based Anomaly Detection), that outperforms AnoGAN's execution time. More recently, Akcay et al. (2018) proposed an approach based on GAN + autoencoders that outperforms EGBAD.

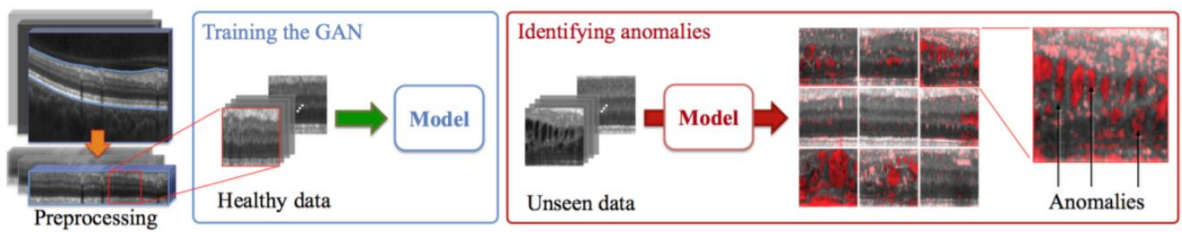

**Figure 2.** GAN generator architecture

A GAN generator G is used to learn the distribution of normal data. In the test, the image finds the appearance of the normal graph through the learned G, and then finds the abnormal or not through comparison.

*2.3. Principles of GAN training model*

The initial significant advancement in GAN training for image generation was the DCGAN architecture proposed by Radford et al. This research further explores the CNN architecture previously used in computer vision and provides guidelines for constructing and training generators and discriminators. In Section III-B, the importance of step length convolution and small step length volume, which are crucial components of architectural design, is mentioned [6-9]. Generators and discriminators can use this technique to improve the quality of image synthesis by learning effective up-sampling and down-sampling operations. To stabilize training in deep models, batch normalization should be used in both networks during training. Additionally, minimizing the number of fully connected layers can improve the feasibility of deep model training. Radford et al. argue that using the leaky ReLU activation function in the discriminator middle layer yields better performance than using the regular ReLU function.

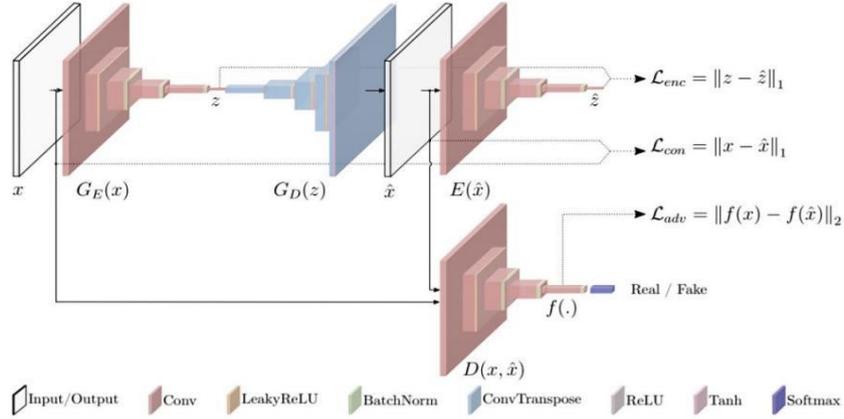

**Figure 2.** GAN adversarial training framework

The adversarial training principle involves training a generator G, constructed from a noise vector Z, to generate a normal data image through several layers of deconvolution. To achieve this, a Gaussian noise vector z is randomly sampled and used to generate a normal image G(z) that corresponds to the test image x, through the trained G[10]. The parameters of G are fixed, and it can only produce images that fall within the normal data distribution. However, it is still necessary to train z as a parameter to be updated and compare the difference between G(z) and x to generate a normal image that closely corresponds to x.

Result 1: If x is a normal image, then x and G(z)
should be the same.

Result 2: If x is abnormal, updating z can reconstruct the ideal normal situation of the abnormal region. This approach enables the identification of abnormal situations and regions by comparing the two figures.

## 3. Methodology

This experiment aims to design and implement an AI fraud prevention system based on Generative Adversarial Network (GANs) algorithm to improve the security of the transaction process. We will effectively protect against fraud by bots or automated systems by creating a collection of adversarial verification graphs to verify that the user in the transaction is a real person. The aim of this experiment is to design and implement a fake face verification code and fraud detection system based on Generative Adversarial network (GANs) algorithm to improve the security of the transaction process. Our focus is on using GAN models to generate adversarial verification graphs to effectively prevent fraud caused by AI face fake captCHA and fictional faces.

Here are the main steps and objectives of the experiment:

### 3.1. Quantitative data establishment

GFP-GAN is based on FFHQ training and consists of 70,000 high-quality images. During training, resize all images to 512². Gfp-gans are trained on synthetic data that approximates real low-quality images and generalizes to real-world images during inference. We follow the convention of using the following degradation model:

$$x = [(y * k_\sigma) \downarrow_r + n_\delta] JPEG_q. \quad (1)$$

① k σ Gaussian fuzzy - High quality image y is first convolved with Gaussian fuzzy kernel k σ, sampling range [0.2:10]
② r - Downsampling based on scale factor r, sampling range [1:8]
③ n δ - Add Gaussian white noise, sampling range [0:15]
④ JPEGq - JPEG image with quality factor q, sampling range [60:100]

It can also be seen from the process of sample construction that GFP-GAN is an unsupervised or self-supervised training model, which does not need manual labeling data.

### 3.2. discriminator model

In defining a discriminator model function define_discriminator(), the input parameter is the size of the picture, the default is (32, 32, 3).

```
1   Model: "sequential"
2   _________________________________________________________________
3   Layer (type)                 Output Shape              Param #
4   =================================================================
5   conv2d (Conv2D)              (None, 32, 32, 64)        1792
6   _________________________________________________________________
7   leaky_re_lu (LeakyReLU)      (None, 32, 32, 64)        0
8   _________________________________________________________________
9   conv2d_1 (Conv2D)            (None, 16, 16, 128)       73856
10  _________________________________________________________________
11  leaky_re_lu_1 (LeakyReLU)    (None, 16, 16, 128)       0
12  _________________________________________________________________
13  conv2d_2 (Conv2D)            (None, 8, 8, 128)         147584
14  _________________________________________________________________
15  leaky_re_lu_2 (LeakyReLU)    (None, 8, 8, 128)         0
16  _________________________________________________________________
17  conv2d_3 (Conv2D)            (None, 4, 4, 256)         295168
18  _________________________________________________________________
19  leaky_re_lu_3 (LeakyReLU)    (None, 4, 4, 256)         0
20  _________________________________________________________________
21  flatten (Flatten)            (None, 4096)              0
22  _________________________________________________________________
23  dropout (Dropout)            (None, 4096)              0
24  _________________________________________________________________
25  dense (Dense)                (None, 1)                 4097
26  =================================================================
27  Total params: 522,497
28  Trainable params: 522,497
29  Non-trainable params: 0
```

After completing the primary data model, it is necessary to normalize the true and false data, in which:
1. Change image pixels from [0,255] to [-1, 1]
2. Randomly take n_sample real pictures from the data set
3. Randomly make n_sample a fake picture
4. Conduct data training on true and false pictures.
The training results are as follows:
After training the truth-false ratio of the truth-false picture, the fake picture needs to be displayed according to the random vector of the generator.
The result of the execution is as follows, because there is no training, so the pictures are gray:

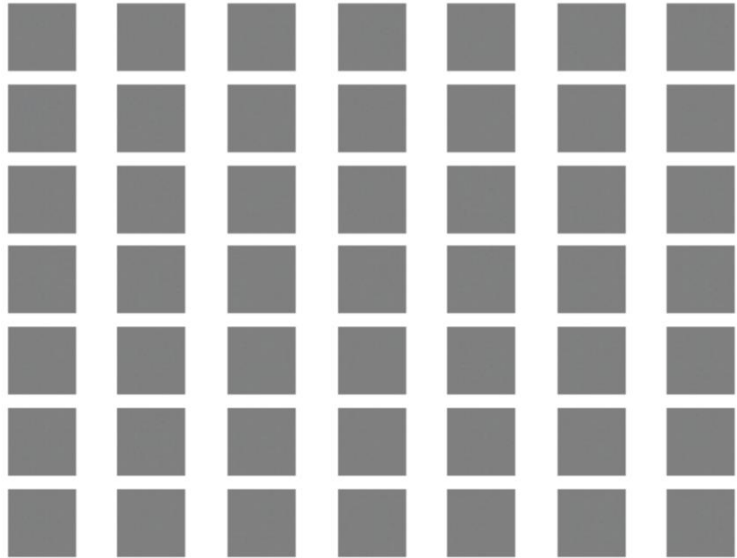

**Figure 3.** Stochastic vector 100-dimensional true-false image primary model

### 3.3. Defining GAN models

- The Gan model is actually composed of Generator and Desciminator
- # define the combined generator and discriminator model, for updating the generator
- def define_gan(g_model, d_model):
-     # make weights in the discriminator not trainable
-     d_model.trainable = False

```
1  Model: "sequential_2"
2  _________________________________________________________________
3  Layer (type)                 Output Shape              Param #
4  =================================================================
5  sequential_1 (Sequential)    (None, 32, 32, 3)         1466115
6  _________________________________________________________________
7  sequential (Sequential)      (None, 1)                 522497
8  =================================================================
9  Total params: 1,988,612
10 Trainable params: 1,466,115
11 Non-trainable params: 522,497
```

### 3.4. The GAN model is trained to display fake images generated by the Generator

Load the trained weight parameters and randomly generate 100 100-dimensional points. Use Generator to generate fake images and display them. The results are as follows:

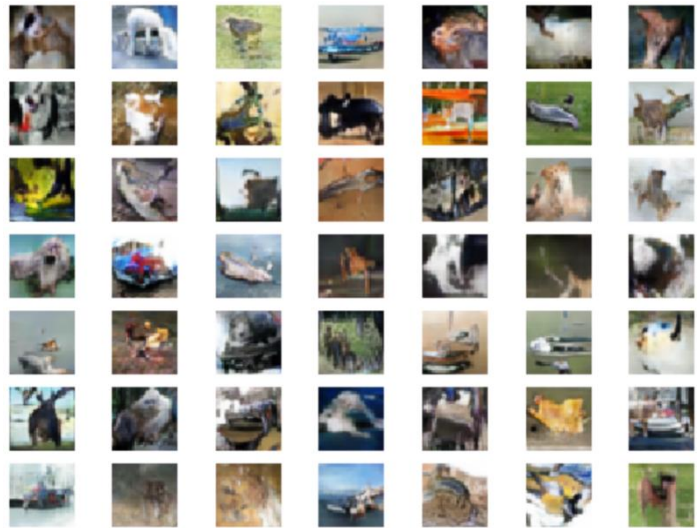

**Figure 4.**10 List of validation diagrams after epoch

After producing a large number of fake pictures/captCHA as well as faces, each image needs to be represented with a vector Z, where this vector Z consists of 100 real numbers in the interval [0,1]. After calculating the distribution of the face image, the generator uses the Gaussian distribution to generate the image from the vector Z. The generator constantly learns to generate new images of faces to fool the discriminator; In turn, the discriminator can better distinguish between the generated face image and the real face image during the competition process.

This project uses GANs to perform vector operations on facial images. In the experiment, the researchers generated images of faces with multiple facial expressions by feeding a series of sample images to the system. For example, it can change a non-smiling face into a smiling face, add an object (such as glasses) to the face, or highlight certain features.

The specific true or false verification face is as follows:

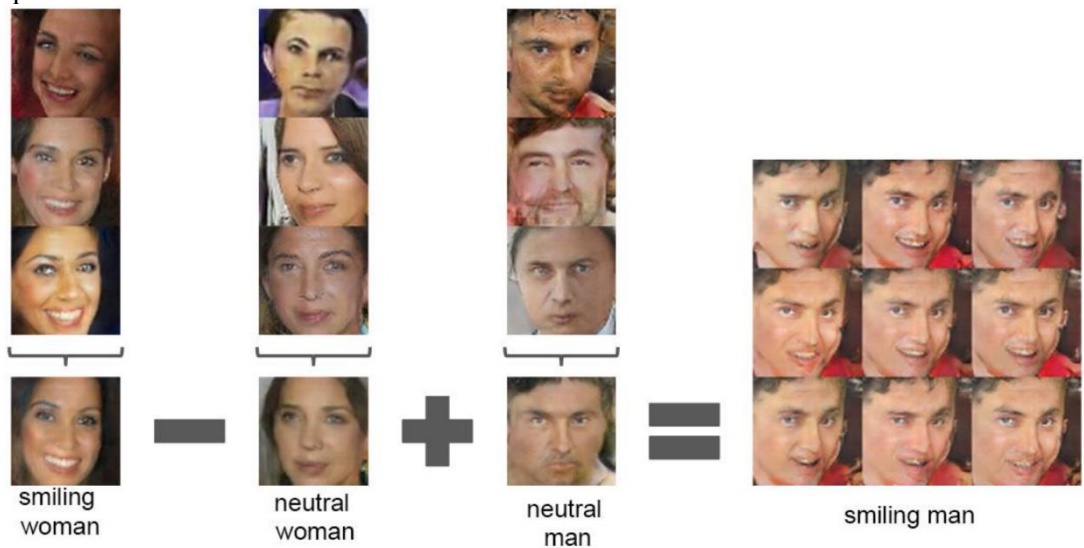

**Figure 5.**Traders verify face display renderings and generate network adversarial validations

In this experiment, the fraud detection of traders in the trading process is distinguished by the form of training the model on the true and false photo sets, and the program of intrusion detection of the adversarial network is generated. The GANs model can also simulate the transformation of face rotation, scaling and shifting, and obtain realistic effects. To achieve this, we first took a series of image samples of the left face and the right face. Then, all the images are averaged to get a Turn

Image Vector. Finally, this "transformation" is applied to new faces by interpolating along the axis of the image vector.

Specifically, we will focus on AI-generated fake captcha for faces and fraud for fictitious faces. By training the GAN model, we aim to improve the accuracy of the system's recognition of these frauds and ensure that the user is a real human being during the transaction, not a fictitious or forged entity, by generating adversarial verification graphs. With this expanded experimental goal, we aim to provide a more comprehensive and reliable solution for preventing face fraud."

In summary, the use of generative adversarial networks (GANs) in fraud detection systems offers significant advantages in performance and accuracy compared to traditional methods. GANs enable the system to better simulate and capture complex patterns of fraudulent behavior, resulting in increased sensitivity to abnormal data. The unsupervised learning capabilities of GANs enable them to identify potentially fraudulent patterns in unlabeled data, thereby enhancing the comprehensiveness and adaptability of detection systems. Furthermore, GANs can generate synthetic data that is more representative, which can aid in expanding the training set of fraud detection systems and improving their ability to identify various fraudulent behaviours. These advantages provide new opportunities for enhanced performance and accuracy in the field of fraud detection.

**4. The challenge of GAN**

GAN technology has made significant progress in the application of images, videos, speech, and natural language. However, there are still some technical challenges that need further study and resolution, such as pattern collapse and pattern collapse issues.

*4.1. Defects in Model Training*

During the training process, GAN is prone to pattern collapse, where the generator can only produce limited types of samples, failing to generate diverse samples. Methods to address this issue include increasing sample diversity, utilizing multiple generators and discriminators to enhance the training algorithm, etc.

*4.2. Training Instability*

The training process of GANs can be complex and prone to issues such as training instability and gradient disappearance. To address these problems, more stable loss functions can be used, the network structure can be improved, and regularization techniques can be employed. Furthermore, challenges related to the lack of training data and data bias may arise [11]. GAN training requires a significant amount of data, and in some application fields, data scarcity or bias may pose challenges. Solutions for improving data include the use of techniques such as data augmentation, transfer learning, and federated learning.

While GAN technology has made significant progress in the application of image, video, speech, and natural language, there are still technical challenges that require further study and resolution, such as pattern collapse and pattern collapse issues.

*4.3. There are defects in model training*

GANs are vulnerable to attacks and interference, leading to unreasonable or dangerous conditions in generated samples. Solutions encompass enhancing the robustness and security of GANs, designing more efficient detection and identification techniques, and tuning parameters securely. Training GANs involves setting multiple hyperparameters, impacting performance and stability. Solutions include designing more efficient automatic parameter tuning technology and improving GAN interpretability.

*4.4. Social and Ethical Issues*

With the widespread use of GANs, social and ethical concerns have arisen. Some possible problems include:

Ethical Issues: GANs, capable of generating realistic fake data, raise ethical concerns related to identity forgery, fraud, etc. Establishing relevant moral and ethical guidelines is necessary to guide GAN use.

Privacy Issues: GANs, generating new data by learning from existing data, may lead to privacy concerns. In the medical field, for instance, GANs can generate lifelike patient data, potentially exposing private information. Relevant privacy protection measures are necessary.

Malicious Use[12]: GANs, capable of generating realistic fake data, may be exploited for malicious purposes like generating fake news and deep forgery. Establishing legal and technical measures is crucial to prevent malicious use.

Fraud and False Information: GANs, generating highly realistic fake data, can create false information, impacting society's credibility and stability. Addressing this issue is a significant social concern.

While many related technologies aim to detect and prevent these problems, they remain significant challenges that GANs currently face. As technology evolves, new issues may arise, emphasizing the need for relevant regulations, guidelines, and ethical standards to govern GAN use and application.

## 5. Conclusion and prospect

### 5.1. conclusion

This study designed and implemented a facial fake verification code and fake face fraud detection system based on the generative adversarial network (GANs) algorithm. The experiments demonstrate that GANs have significant advantages in fraud detection compared to traditional methods. Through the use of an adversarial training mechanism, it is possible to simulate normal data distribution more accurately. This can effectively prevent fraud caused by fake face verification codes and fictitious faces.

### 5.2. The importance of experimental verification

The research emphasizes the importance of effectively verifying user authenticity by generating adversarial verification graphs[13]. Our experimental results demonstrate the critical role of this mechanism in the transaction process, especially against fraud caused by fake captCHA and fictitious faces generated by AI. The introduction of this kind of verification graph effectively improves the security of transactions, and has substantial significance for preventing the fraud of robots and automated systems.

### 5.3. The future of GANs fraud detection

Future research will aim to improve the accuracy of generative models and optimize generative adversarial networks to combat new fraud methods. Multimodal data, such as voice and behavioral patterns, could be introduced to build a more comprehensive fraud detection system. Additionally, efforts will be made to apply the algorithm to real-time trading environments and ensure the system can scale to adapt to growing data and trading volume. Ultimately, our focus is on improving the interpretability of generative adversarial networks to ensure that users and decision-makers can understand the decision-making process of the model. This will enhance the credibility of the overall system.

**Acknowledgments**
During the completion of this study, I would like to sincerely thank Liu, Bo and others for their outstanding contributions to the academic community. Their journal article "Integration and Performance Analysis of Artificial Intelligence and Computer Vision Based on Deep Learning. Algorithms provides valuable inspiration and enlightenment for my research on financial portfolio management and risk prediction in the field of machine reinforcement learning. This paper deeply discusses the integration and performance analysis of artificial intelligence and computer vision based on deep learning algorithms, providing me with rich theoretical support in experimental design and reinforcement learning application. Some of the core conclusions and methods adopted in my research are derived from this article, which has played an important guiding role for my research. I would like to express my sincere gratitude to Liu, Bo and others for bringing outstanding research to the academic community and providing valuable reference resources for my research.

## 6. References

[1] Aftabi Seyyede ZahraAhmadi Ali,and Farzi Saeed. "Fraud detection in financial statements using data mining and GAN models." Expert Systems With Applications 227. (2023):

[2] Strelcenia Emilija,and Prakoonwit Simant. "A Survey on GAN Techniques for Data Augmentation to Address the Imbalanced Data Issues in Credit Card Fraud Detection." Machine Learning and Knowledge Extraction 5. 1 (2023):

[3] Jeonghyun Hwang,and Kangseok Kim. "An Efficient Domain-Adaptation Method using GAN for Fraud Detection." International Journal of Advanced Computer Science and Applications (IJACSA) 11. 11 (2020):

[4] Yue YangChenyuan Liu,and Ningning Liu. "Credit Card Fraud Detection based on CSat-Related AdaBoost". Ed. 2019.

[5] Tianbo, Song, Hu Weijun, Cai Jiangfeng, Liu Weijia, Yuan Quan, and He Kun. "Bio-inspired Swarm Intelligence: a Flocking Project With Group Object Recognition." In 2023 3rd International Conference on Consumer Electronics and Computer Engineering (ICCECE), pp. 834-837. IEEE, 2023.DOI: 10.1109/mce.2022.3206678

[6] Liu, B., Zhao, X., Hu, H., Lin, Q., & Huang, J. (2023). Detection of Esophageal Cancer Lesions Based on CBAM Faster R-CNN. Journal of Theory and Practice of Engineering Science, 3(12), 36–42. https://doi.org/10.53469/jtpes.2023.03(12).06

[7] Liu, Bo, et al. "Integration and Performance Analysis of Artificial Intelligence and Computer Vision Based on Deep Learning Algorithms." arXiv preprint arXiv:2312.12872 (2023).

[8] Yu, L., Liu, B., Lin, Q., Zhao, X., & Che, C. (2024). Semantic Similarity Matching for Patent Documents Using Ensemble BERT-related Model and Novel Text Processing Method. arXiv preprint arXiv:2401.06782.

[9] Zhao, Xinyu, et al. "Effective Combination of 3D-DenseNet's Artificial Intelligence Technology and Gallbladder Cancer Diagnosis Model". Frontiers in Computing and Intelligent Systems, vol. 6, no. 3, Jan. 2024, pp. 81-84, https://doi.org/10.54097/iMKyFavE.

[10] Li, Shulin, et al. "Application Analysis of AI Technology Combined With Spiral CT Scanning in Early Lung Cancer Screening". Frontiers in Computing and Intelligent Systems, vol. 6, no. 3, Jan. 2024, pp. 52-55, https://doi.org/10.54097/LAwfJzEA.

[11] Xiong Li, et al. "Improving Robot-Assisted Virtual Teaching Using Transformers, GANs, and Computer Vision." *Journal of Organizational and End User Computing (JOEUC) 36.* 1 (2024):

[12] Park KeundeokErgan Semiha,and Feng Chen. "Quality assessment of residential layout designs generated by relational Generative Adversarial Networks (GANs)." *Automation in Construction 158.* (2024):

[13] Meng Xianglong, et al. "DedustGAN: Unpaired learning for image dedusting based on Retinex with GANs." *Expert Systems With Applications 243.* (2024):